\documentclass[runningheads]{llncs}
\usepackage[T1]{fontenc}
\usepackage{graphicx}
\usepackage{bm}
\usepackage{hyperref}
\usepackage{amsmath}
\usepackage{amssymb}
\usepackage{mathtools}
\usepackage[ruled, linesnumbered]{algorithm2e}
\SetKwInOut{Input}{Input}
\SetKwInOut{Return}{Return}

\newcommand{\R}{\mathbb{R}}
\newcommand{\norm}[1]{\left\lVert#1\right\rVert}
\begin{document}
\title{Deep Gaussian mixture model for unsupervised image segmentation}
%
%\titlerunning{Segmentation with deep learning and Gaussian mixture models}
% If the paper title is too long for the running head, you can set
% an abbreviated paper title here
%
\author{Matthias Schwab\inst{1,2} \and
Agnes Mayr \inst{1} \and Markus Haltmeier\inst{2}}
\authorrunning{M. Schwab et al.}
% First names are abbreviated in the running head.
% If there are more than two authors, 'et al.' is used.
%
\institute{Medical University of Innsbruck, Innsbruck, Tirol, Austria 
\email{matthias.schwab@i-med.ac.at}\and
University of Innsbruck, Innsbruck, Tirol, Austria}
\maketitle              % typeset the header of the contribution
\begin{abstract}
The recent emergence of deep learning has led to a great deal of work on designing supervised deep semantic segmentation algorithms. As in many tasks sufficient pixel-level labels are very difficult to obtain, we propose a method which combines a Gaussian mixture model (GMM) with unsupervised deep learning techniques. In the standard GMM the pixel values with each sub-region are modelled by a Gaussian distribution. In order to identify the different regions, the parameter vector that minimizes the negative log-likelihood (NLL) function regarding the GMM has to be approximated. For this task, usually iterative optimization methods such as the expectation-maximization (EM) algorithm are used. In this paper, we propose to estimate these parameters directly from the image using a convolutional neural network (CNN). We thus change the iterative procedure in the EM algorithm replacing the expectation-step by a gradient-step with regard to the networks parameters. This means that the network is trained to minimize the NLL function of the GMM which comes with at least two advantages. As once trained, the network is able to predict label probabilities very quickly compared with time consuming iterative optimization methods. Secondly, due to the deep image prior our method is able to partially overcome one of the main disadvantages of GMM, which is not taking into account correlation between neighboring pixels, as it assumes independence between them. We demonstrate the advantages of our method in various experiments on the example of myocardial infarct segmentation on multi-sequence MRI images.

\end{abstract}

\section{Introduction}

Image segmentation is the process of partitioning a given image into semantically meaningful regions, by assigning a label to each pixel. It plays a crucial role in various applications as for example medical image analysis \cite{ramesh2021review}, or autonomous driving \cite{papadeas2021real}. Traditional segmentation techniques relied on hand-crafted heuristics and manually tuned parameters to extract visual features from the images \cite{comaniciu2002mean,arbeláez2011contour}. However, these methods may not generalize well across different domains and imaging conditions. In recent years, due to the ability of neural networks to learn how to extract information from their input, deep learning has emerged as a powerful paradigm for image segmentation. Because of the rapid growth of open-source datasets containing millions of annotated images across various object categories \cite{lin2014microsoft,kuznetsova2020open,cordts2016cityscapes}, neural networks could be trained in a supervised manner to produce highly accurate segmentation masks. 

However, acquiring labeled data for segmentation tasks can be quite time expensive and labor-intensive, especially for applications involving complex multi-modal images. To this end unsupervised image segmentation methods offer a promising alternative by using the intrinsic structure and statistics of images to generate segmentation masks without the need for labeled data. Often combined with deep learning techniques, these methods typically cluster pixels or regions based on similarities in color, texture, or other low-level features, thereby partitioning the image into reasonable segments \cite{gruber2023variational,xia2017w}.

One of the methods used for unsupervised image segmentations are Gaussian mixture models (GMM). These models are based on the assumption that the intensity (gray scale or color) value of each pixel in a given image is a sample from a mixture of Gaussian distributions. After estimating the optimal parameters of this distributions a suitable labeling rule is applied to assign labels to the pixels of an image. As this segmentation technique has shown to be quite useful in some medical image segmentation applications \cite{balafar2014gaussian,liu2014myocardium}, its main weakness is that it is only based on pixel intensities, not taking into account any spatial information. Therefore, GMMs have been combined with spatial regularization techniques using probabilistic atlases \cite{zhuang2018multivariate}, Markov random fields \cite{rajapakse1997statistical}, or coupled  level sets \cite{liu2017myocardium}. Even though these regularization techniques often provide some improvements for the final segmentation performance, the resulting algorithms have been found to be substantially more complex and computationally expensive. 
%[maybe [19]]

In this paper we present two novel segmentation methods which we call deepG and deepSVG. These methods combine the useful properties of deep learning  with the classical GMM approach. To be precise, we are incorporating a convolutional neural network (CNN) in the optimization procedure of estimating the optimal parameters of the GMM. This comes with at least two  advantages: 
\begin{itemize}
    \item[$\bullet$] Due to the convolution-based architecture of the neural network (deep image prior) deepG is able to partly overcome the main disadvantage of the GMM, which does not take into account the correlation between neighboring pixels.
    \item[$\bullet$] As once trained, the neural network is able to predict label probabilities very quickly compared with the usually used time consuming iterative optimization methods.
\end{itemize}

The rest of the paper is organized as follows. Section \ref{sec:gmm} describes
the GMM and a more flexible method called spatially variant GMM (SVGMM) in full detail. In Section \ref{sec:method}, the proposed methods deepG and deppSVG are introduced. Section \ref{sec:experiments} shows numerical experiments on the real life problem of myocardial image segmentation. Finally, Section \ref{sec:conclusion} gives a short summary of the presented research and discusses possible future work.

\section{Gaussian mixture model}
\label{sec:gmm}

Let $\boldsymbol{I} = \{I_1, \dots, I_m \}$ be a set of $m$ images acquired from the same object. For $m>1$, $\boldsymbol{I}$ could be an RGB-image with different color channels or spatially aligned medical images from multiple imaging modalities. In the further course of the work, we will refer to this set of $m$ images of the same object simply as an image. We denote the spatial domain of the images by $\Omega \subset \R^2$. Then $\boldsymbol{I}(x) \in \R^m$ denotes the $m$ different observations at location $x \in \Omega$. Let $K \subset \mathbb{N}$ be the finite set of labels or classes one wants to divide the images into. Further, let $s(x) \in K$ denote the class to which the pixel at location $x$ belongs to. As GMMs assume the intensity distributions of the classes to be Gaussian, the probability density function of a class $k \in K$ is given by
\begin{align*}
    p(\boldsymbol{I}(x)|s(x)=k, \mu_k, \Sigma_k)= g(\boldsymbol{I}(x) |\mu_k, \Sigma_k),
\end{align*}
where $g$ is an $m$-variate Gaussian density function of the form 
\begin{align*}
    g(\boldsymbol{I}(x) |\mu_k, \Sigma_k) = \frac{1}{(2 \pi)^{m/2} \det(\Sigma_k)^{1/2}} \exp \left(-\frac{1}{2} (\boldsymbol{I}(x)- \mu_k)^T \Sigma_k^{-1} (\boldsymbol{I}(x)- \mu_k )\right),
\end{align*}
with mean vector $\mu_k \in \R^m$ and covariance matrix $\Sigma_k \in \R^{m \times m}$. By using the different label proportions $p(k) =\pi_k$, which describe the probabilities of the different classes, the GMM defines the density function of the observation $\boldsymbol{I}(x)$ as 
\begin{align*}
    p(\boldsymbol{I}(x)| \boldsymbol{\pi}, \boldsymbol{\mu}, \boldsymbol{\Sigma}) &= \sum_{k\in K} p(k) \; p(\boldsymbol{I}(x)|s(x)=k, \mu_k, \Sigma_k)\\
    &= \sum_{k\in K} \pi_k g(\boldsymbol{I}(x) |\mu_k, \Sigma_k), 
\end{align*}
with mixing weights $\boldsymbol{\pi} = (\pi_k| k \in K) \in  \R^{|K|}$, means $\boldsymbol{\mu}= (\mu_k| k \in K) \in \R^{|K| \times m}$ and covariance matrices $\boldsymbol{\Sigma} = (\Sigma_k | k \in K) \in \R^{|K| \times m \times m}$.  If it is assumed that the pixel intensities are independent of their location in the image, the joint conditional density of the GMM is given by
\begin{align}
\label{eq:LH}
    p(\boldsymbol{I}|\boldsymbol{\pi}, \boldsymbol{\mu}, \boldsymbol{\Sigma}) = \prod_{x \in \Omega} \sum_{k\in K} \pi_k g(\boldsymbol{I}(x) |\mu_k, \Sigma_k). 
\end{align}
Given an image $\boldsymbol{I}$ the goal is to find an estimate of the parameters $(\boldsymbol{\pi}, \boldsymbol{\mu}, \boldsymbol{\Sigma})$ that maximize the likelihood in \eqref{eq:LH}, which is equivalent to minimizing the negative log-likelihood (NLL) given by
\begin{align}
\label{eq:NLL}
    \operatorname{NLL}(\boldsymbol{I} | \boldsymbol{\pi}, \boldsymbol{\mu}, \boldsymbol{\Sigma}) = -\frac{1}{|\Omega|}\sum_{x \in \Omega} \log \left( \sum_{k\in K} \pi_k g(\boldsymbol{I}(x) |\mu_k, \Sigma_k) \right).
\end{align}
In order to estimate mixing weights, mean vectors and covariance matrices that minimize (\ref{eq:NLL}) usually the expectation-maximization (EM) algorithm \cite{dempster1977maximum} is used. This algorithm is shown in Algorithm \ref{alg:EM-GMM} and consists of two major steps: An expectation-step (E-step) followed by a maximization-step (M-step). In the case of maximum likelihood (ML) estimation for the GMM the E-step corresponds to calculating the probabilities for each pixel belonging to the different classes, given the current parameters and the pixel values at position $x$. The M-step then uses the probabilities calculated in the E-step to obtain a new estimate of the parameters which reduce the NLL. These two steps are then iterated until convergence. 
\begin{algorithm}[t]
\caption{EM Algorithm for GMM}\label{alg:EM-GMM}
\KwIn{Image $\boldsymbol{I}$ and set of classes $K$}
\BlankLine
Initialize parameters $(\boldsymbol{\pi}, \boldsymbol{\mu}, \boldsymbol{\Sigma}) \in \R^{|K|} \times \R^{|K| \times m} \times \R^{|K| \times m \times m}$\;
\Repeat{convergence}
        {\textbf{E-step:} For all $x \in \Omega$ and all $k \in K$ set
        \begin{align*}
            w_{xk} \leftarrow \frac{\pi_k  g(\boldsymbol{I}(x) |\mu_k , \Sigma_k ) }
                                {\sum_{k' \in K} \pi_{k'}  g\left(\boldsymbol{I}(x) |\mu_{k'} , \Sigma_{k'} \right)}
        \end{align*}
        
        \noindent \textbf{M-step:} For all $k \in K$ set
        \begin{align*}
        \pi_k &\leftarrow \frac{1}{|\Omega|} \sum_{x \in \Omega} w_{xk}  \\
        \mu_k &\leftarrow \frac{\sum_{x \in \Omega} w_{xk}  \boldsymbol{I}(x)}{\sum_{x \in \Omega} w_{xk} } \\
        \Sigma_k &\leftarrow \frac{\sum_{x \in \Omega} w_{xk}  (\boldsymbol{I}(x)-\mu_k ) (\boldsymbol{I}(x)-\mu_k )^T}{\sum_{x \in \Omega} w_{xk} }
        \end{align*}
        }
    \Return{$\boldsymbol{\pi} , \boldsymbol{\mu} , \boldsymbol{\Sigma}$}
\end{algorithm}
As shown in \cite{titterington1985statistical} the likelihood function of \eqref{eq:LH} exhibits local maxima and the EM algorithm will converge to one of them. As the parameters $\boldsymbol{\pi} , \boldsymbol{\mu} , \boldsymbol{\Sigma} $ which are returned by Algorithm \ref{alg:EM-GMM} describe global properties of the measurements in the image $\boldsymbol{I}$, they do not tell how to assign labels to pixels. Therefore, a suitable labeling rule is needed to determine the final segmentation mask. For this task usually an additional E-step is performed, calculating the probabilities for each pixel belonging to the different classes given the estimated parameters. These probabilities can be calculated using Bayes rule
%\begin{align*}
%    p_{xk} = \frac{\pi_k  g(\boldsymbol{I}(x) |\mu_k , \Sigma_k ) }
%                                {\sum_{k' \in K} \pi_{k'}  g\left(\boldsymbol{I}(x) |\mu_{k'} , %\Sigma_{k'} \right)}.
%\end{align*}
\begin{align*}
    w_{xk} &= p(s(x)=k|\boldsymbol{I}(x), \boldsymbol{\pi},\boldsymbol{\mu}, \boldsymbol{\Sigma}) \\
    &= \frac{p(s(x)=k|\boldsymbol{\pi},\boldsymbol{\mu}, \boldsymbol{\Sigma}) p(\boldsymbol{I}(x)|s(x)=k, \boldsymbol{\pi},\boldsymbol{\mu}, \boldsymbol{\Sigma})}{p(\boldsymbol{I}(x)|\boldsymbol{\pi},\boldsymbol{\mu}, \boldsymbol{\Sigma})}\\ 
    &=\frac{\pi_k  g(\boldsymbol{I}(x) |\mu_k , \Sigma_k ) }
                                {\sum_{k' \in K} \pi_{k'}  g\left(\boldsymbol{I}(x) |\mu_{k'} , \Sigma_{k'} \right)}.
\end{align*}
Then for each pixel the class with the highest probability is assigned as final label.  

\subsection{Spatially Variant GMM}

\begin{algorithm}[!b]
\caption{EM Algorithm for SVGMM}\label{alg:EM-SVGMM}
\KwIn{Image $\boldsymbol{I}$ and set of classes $K$}
\BlankLine
Initialize parameters $(\boldsymbol{\Pi}, \boldsymbol{\mu}, \boldsymbol{\Sigma}) \in \R^{|K| \times |\Omega|} \times \R^{|K| \times m} \times \R^{|K| \times m \times m}$\;
\Repeat{convergence}{
    \textbf{E-step:} For all $x \in \Omega$ and all $k \in K$ set
    \begin{align*}
            w_{xk} \leftarrow \frac{\Pi_{k}(x) g(\boldsymbol{I}(x) |\mu_k, \Sigma_k)}
                                {\sum_{k' \in K} \Pi_{k'}(x) g\left(\boldsymbol{I}(x) |\mu_{k'}, \Sigma_{k'}\right)}
    \end{align*}
    
    \textbf{M-step:} For all $x \in \Omega $ and all $k \in K$ set
        \begin{align*}
        \Pi_{k}(x) &\leftarrow \frac{w_{xk}}{\sum_{k' \in K} w_{xk'}} = w_{xk} \\
        \mu_k  &\leftarrow \frac{\sum_{x \in \Omega} w_{xk} \boldsymbol{I}(x)}{\sum_{x \in \Omega} w_{xk}} \\
        \Sigma_k &\leftarrow \frac{\sum_{x \in \Omega} w_{xk} (\boldsymbol{I}(x)-\mu_k) (\boldsymbol{I}(x)-\mu_k)^T}{\sum_{x \in \Omega} w_{xk}}
        \end{align*}
        }
    \Return{$\boldsymbol{\Pi}, \boldsymbol{\mu}, \boldsymbol{\Sigma}$}

\end{algorithm}
In \cite{sanjay1998bayesian} it is pointed out that the final labeling rule for the GMM is not optimal as the pixel labels are computed on the basis of ML estimates for $\boldsymbol{\pi}, \boldsymbol{\mu}$ and $\boldsymbol{\Sigma}$, but are not ML estimates themselves. Therefore, with the goal of developing a more flexible model for pixel labeling, the spatially variant Gaussian mixture model (SVGMM) was introduced \cite{sanjay1998bayesian}. The only difference compared to the classical GMM is that in the SVGMM the different label proportions are assigned for each pixel individually. So for each location $x \in \Omega$, $p(s(x)=k)=\Pi_k(x)$ denotes the probability of this pixel belonging to class $k \in K$. Given an image $\boldsymbol{I}$ the SVGMM then defines the NLL function 
\begin{align*}
    \operatorname{NLL_V}(\boldsymbol{I} | \boldsymbol{\Pi}, \boldsymbol{\mu}, \boldsymbol{\Sigma}) = -\frac{1}{|\Omega|}\sum_{x \in \Omega} \log \left( \sum_{k\in K} \Pi_{k}(x) g(\boldsymbol{I}(x) |\mu_k, \Sigma_k) \right),
\end{align*}
where $\boldsymbol{\Pi} = (\Pi_k|k \in K)\in \R^{|K|} \times \R^{|\Omega|}$ and $\sum_{k \in K} \Pi_{k}(x) = 1$ for all $x \in \Omega$. The classical GMM can be seen as a special case of the SVGMM with $\Pi_{k}(x) \equiv \pi_k$ being constant over all $x \in \Omega$. Algorithm \ref{alg:EM-SVGMM} shows the EM algorithm for ML estimation of the SVGMM. Also in \cite{sanjay1998bayesian} it is shown that under some mild conditions the label probabilities $\Pi_{k}(x)$ converge to zeros and ones, so that the final labeling step is not necessary for the SVGMM with output $\boldsymbol{\Pi}$ already containing the desired binary segmentation masks. In our numerical experiments in Section \ref{sec:experiments} , we were able to observe this advantage of the SVGMM very clearly.

\section{Proposed deep Gaussian mixture models}
\label{sec:method}

One can observe that in the E-steps of Algorithm \ref{alg:EM-GMM} and Algorithm \ref{alg:EM-SVGMM} probabilities $w_{xk}$ are calculated for each pixel individually. Also the calculated values $w_{xk}$ only depend on the image intensity at position $x$ and not on any other intensity values of neighboring pixels. Therefore, we suggest not to calculate the weights $\boldsymbol{w}= (w_{xk}| x\in \Omega, k \in K)$ all individually as in the conventional E-step, but to predict them all together directly from the whole input image $\boldsymbol{I}$. For this we use a CNN $\phi$ with U-net architecture \cite{ronneberger2015u} and learnable parameters $\theta$. The update in the E-step is then replaced by an update of the network's parameters $\theta$, executing a Gradient-step with respect to the NLL function. The new weights $\boldsymbol{w}$ are then defined to be the output of the updated network. After each Gradient-step a conventional M-step is done to update the parameters of the mixture model. These two steps are then iterated until convergence. By either choosing model parameters and updating rules in Algorithm \ref{alg:EM-GMM} or Algorithm \ref{alg:EM-SVGMM}, our proposed method can be applied to both, the classical GMM and the SVGMM. We refer to these two resulting methods as deepG and deepSVG. The deepSVG method is fully described in Algortihm \ref{alg:proposed}. 
\begin{algorithm}[!b]
\caption{Proposed algorithm (deepSVG).}
\label{alg:proposed} 
\KwIn{Image $\boldsymbol{I}$ and set of classes $K$}
\BlankLine
Initialize weights $\theta$ of the CNN $\phi$ and choose learning rate $\alpha$\;
Set $\boldsymbol{w} \leftarrow \phi_\theta(\boldsymbol{I})$\;
Do an M-step as in Algorithm \ref{alg:EM-SVGMM} to obtain $(\boldsymbol{\Pi}, \boldsymbol{\mu}, \boldsymbol{\Sigma})$\;
\Repeat{convergence}{
    \textbf{Gradient step:} Set
            \begin{align*}
                \theta &\leftarrow \theta - \alpha \nabla_\theta \operatorname{NLL_V}(\boldsymbol{I} | \boldsymbol{\Pi}, \boldsymbol{\mu}, \boldsymbol{\Sigma}) \\
                \boldsymbol{w} &\leftarrow \phi_\theta(\boldsymbol{I})
            \end{align*}
            
            \textbf{M-step} as in Algorithm \ref{alg:EM-SVGMM} to update $(\boldsymbol{\Pi}, \boldsymbol{\mu}, \boldsymbol{\Sigma})$
        }
    \Return{$\boldsymbol{\Pi}, \boldsymbol{\mu}, \boldsymbol{\Sigma}$}
\end{algorithm}

The advantage of our proposed methods is that, due to their convolutional architecture, the network's outputs at a position $x$ not only depend on the image intensities at that exact position, but also on the intensities of neighboring pixels. This deep image prior thus implicitly incorporates spatial regularization into the method. The proposed algorithm is also very flexible for incorporating further regularization to the desired outputs. This can be achieved by simply adding a regularization term to the loss function of the network. The regularized algorithm then is exactly the same as Algorithm \ref{alg:proposed} except that the Gradient-step for $ \operatorname{NLL_V}(\boldsymbol{I} | \boldsymbol{\Pi}, \boldsymbol{\mu}, \boldsymbol{\Sigma})$ is replaced by a Gradient-step for 
\begin{align*}
    \operatorname{NLL_V}(\boldsymbol{I} | \boldsymbol{\Pi}, \boldsymbol{\mu}, \boldsymbol{\Sigma}) +\lambda r(\boldsymbol{\Pi}, \boldsymbol{\mu}, \boldsymbol{\Sigma}),
\end{align*}
with regularization parameter $\lambda \in \R$ and regularizing function $r$.

Usually in deep learning one trains a neural network on a training dataset consisting of multiple images and expects that the trained model will also work on unseen images of a similar structure. This comes with the advantage that once a neural network has been trained on a large dataset, it can also be applied to new images making optimization for each individual image unnecessary. However, for this to work it is necessary to have access to a dataset that contains enough images to represent a large enough sample of the overall set of images on which the final method should work. Of course, if a such a  training dataset is available, our proposed method could also be used to train a single network on the mulitple images of the training dataset. The proposed training procedure, which can also include potential regularization techniques is described in Algorithm \ref{alg:proposed_train}. After training, the returned network can then also be applied directly to new images.

\begin{algorithm}[t]
\caption{Training on multiple images with potential regularization.}
\label{alg:proposed_train}   
 \KwIn{Training images $\{\boldsymbol{I}_1, \dots \boldsymbol{I}_N \}$ and set of classes $K$}
\BlankLine
Initialize weights $\theta$ of the CNN $\phi$, choose learning rate $\alpha$, regularization parameter $\lambda$ and regularizing function $r$\;
\For{$i=1, \dots, N$} 
    {Set $\boldsymbol{w}_i \leftarrow \phi_\theta(\boldsymbol{I}_i)$\;
    Do an M-step as in Algorithm \ref{alg:EM-SVGMM} to obtain $(\boldsymbol{\Pi}_i, \boldsymbol{\mu}_i, \boldsymbol{\Sigma}_i)$\;}
    
   \Repeat{convergence}{
        \textbf{Gradient-step:} Set
            \begin{align*}
                \theta \leftarrow \theta - \alpha \nabla_\theta \sum_{i=1}^N \operatorname{NLL_V}(\boldsymbol{I}_i | \boldsymbol{\Pi}_i, \boldsymbol{\mu}_i, \boldsymbol{\Sigma}_i) +\lambda r(\boldsymbol{\Pi}_i, \boldsymbol{\mu}_i, \boldsymbol{\Sigma}_i)
            \end{align*}
        
          \For{$i=1, \dots, N$}
            {Set $\boldsymbol{w}_i \leftarrow \phi_\theta(\boldsymbol{I}_i)$\;
            Perform an \textbf{M-step} as in Algorithm \ref{alg:EM-SVGMM} to update $(\boldsymbol{\Pi}_i, \boldsymbol{\mu}_i, \boldsymbol{\Sigma}_i)$}}

\Return{$\phi_\theta$}
\end{algorithm}

\section{Experiments and results}
\label{sec:experiments}

\begin{figure}[t]
\includegraphics[width=\textwidth]{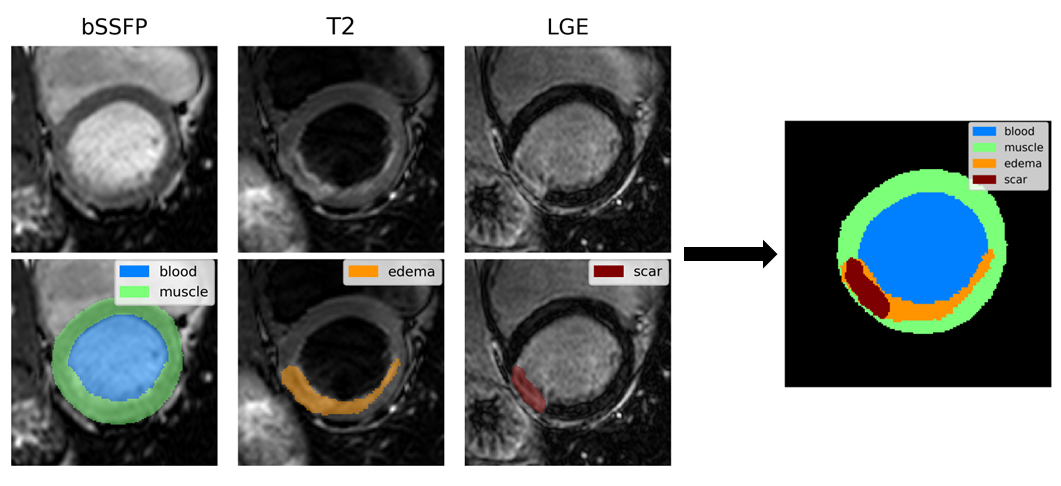}
\caption{Visualisation of myocardial pathology segmentation by combining three sequences of CMR images acquired from the same patient. The bSSFP sequence is used to identify myocardial borders. Edema is marked in the T2-weighted CMR image and scar is quantified in the LGE image.} \label{fig:1}
\end{figure}

We tested our method on a public dataset \cite{zhuang2018multivariate} consisting of $25$ paired three-sequence cardiac magnetic resonance (CMR) images obtained from patients after acute myocardial infarction. These include  balanced steady-state free precession (bSSFP), late gadolinium enhancement (LGE) and T2-weighted CMR sequences in short axis orientation, which were preprocessed to spatially align all the different sequences. As bSSFP images can be used to detect the myocardial borders, T2-weighted CMR can display myocardial edema after acute myocardial infarction. Furthermore, in LGE images, the position and distribution of myocardial infarct (also known as scar) can be quantified. By combining the information of the three-sequence CMR images, the tissue of the left ventricle can be divided into four classes: Blood pool, healthy myocardium, myocardial edema and myocardial scar. On the $25$ patients of the dataset manual labeling of these four tissue types was performed slice-by-slice as illustrated in Figure~\ref{fig:1}.

\subsection{Optimizing on single images}
We tested how the different algorithms described earlier are able to divide the left ventricle into the four different tissue types. To only extract the pixels of the left ventricle, we used manual labeling of the left ventricular borders. As the mixture models need the number of classes one wants to divide the image into to be known in advance, we only included images of the dataset which contained all the four tissue types. This resulted in $79$ multi-sequence CMR images on which we tested four different methods:
\begin{itemize}
    \item[$\bullet$] Algorithm \ref{alg:EM-GMM} (GMM)
    \item[$\bullet$] Algorithm \ref{alg:proposed} adapted to the classical GMM (deepG)
    \item[$\bullet$] Algorithm \ref{alg:EM-SVGMM} (SVGMM)
    \item[$\bullet$] Algorithm \ref{alg:proposed} (deepSVG)
\end{itemize}
\begin{figure}[t]
\includegraphics[width=\textwidth]{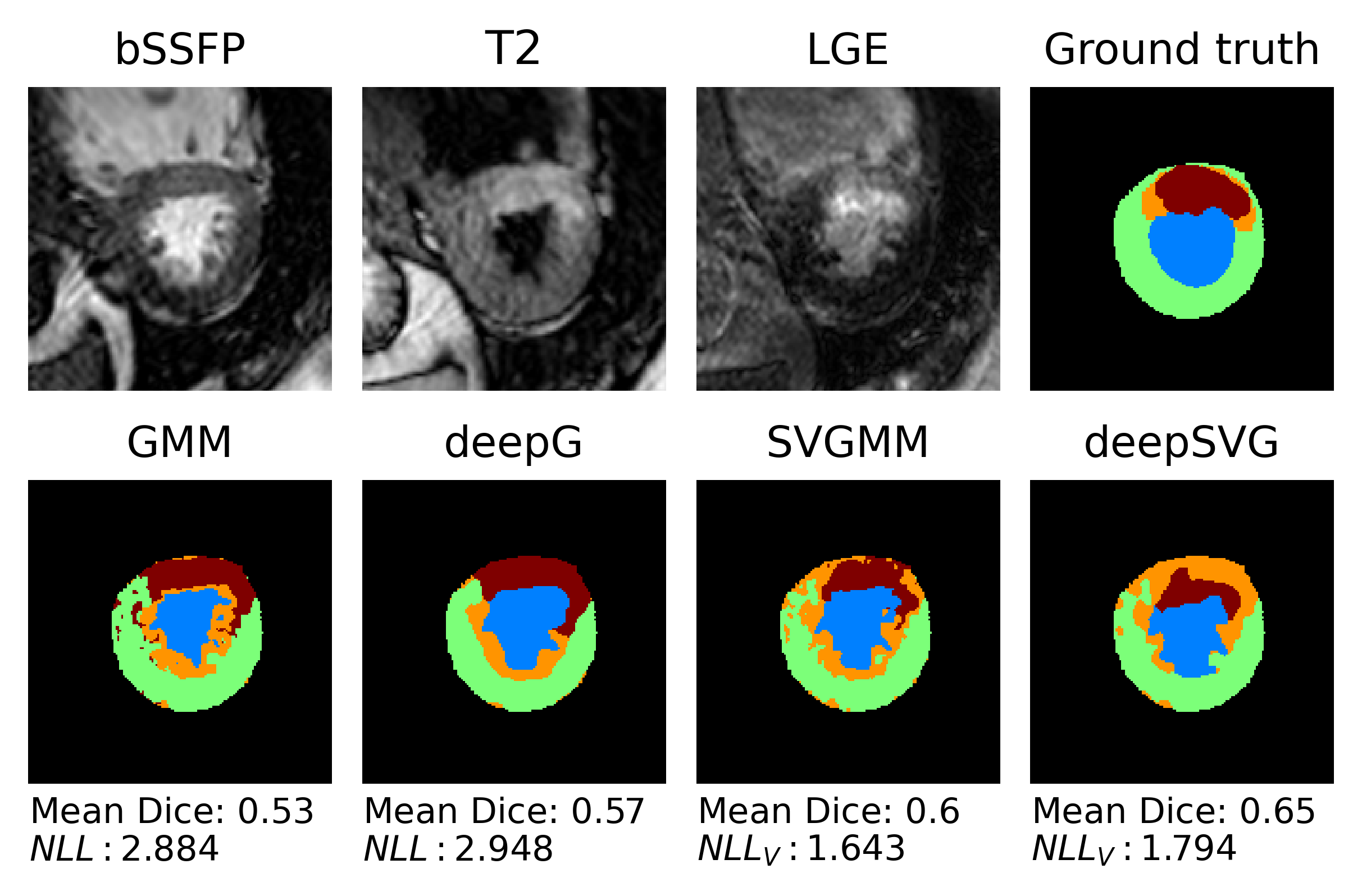}
\caption{Example of a three-sequence CMR image with according ground truth segmentation (top) and segmentation masks produced by the different methods (bottom). The CNN-based methods (deepG and deepSVG) exhibit slightly higher values for the NLLs, but their segmentations are smoother resulting in better Dice scores.}
\label{fig:2}
\end{figure}

\begingroup
\setlength{\tabcolsep}{5pt}
\begin{table}[!b]
\caption{A comparison of Dice coefficient of the different methods for the four different tissue types. The entries show the mean values $\pm$ standard deviations obtained over all $79$ images. The best scores are marked in bold.}\label{tab1}
\centering
\begin{tabular}{l| c c c c c c}
\hline 
Method  &  blood               & myocardium           & edema                & scar                 & mean       \\
\hline \hline
GMM     & $0.75 \pm 0.14$      & $0.73 \pm 0.13$      & $0.13 \pm 0.19$      & $0.38 \pm 0.22$      & $0.50 \pm 0.04$    \\
deepG   & $0.78 \pm 0.12$      & $\bm{0.77 \pm 0.13}$      & $0.08 \pm 0.16$ & $0.48 \pm 0.20$      & $0.53 \pm 0.03$    \\
SVGMM   & $0.77 \pm 0.08$      & $0.74 \pm 0.10$      & $\bm{0.18 \pm 0.19}$ & $0.49 \pm 0.17$      & $0.55 \pm 0.05$  \\
deepSVG & $\bm{0.79 \pm 0.07}$ & $\bm{0.77 \pm 0.09}$ & $0.15 \pm 0.18$      & $\bm{0.52 \pm 0.17}$ & $\bm{0.56 \pm 0.05}$  \\
\hline
\end{tabular}
\end{table}
\endgroup

For all the algorithms we initialized randomly and stopped when the NLL gain per iteration went below a threshold of $0.001$. Since ground truth labels were available for this dataset, we measured segmentation accuracy using the Dice coefficient
\begin{equation}
    \label{eq:dice}
    \mathrm{DICE} = \frac{2 |P \cap G|}{|P|+|G|},
\end{equation}
where $P$ presents the segmentation prediction of the tested method and $G$ the manual gold standard. As all the methods work unsupervised and initialization was done randomly, the label values, which the methods assigned to the different classes, were also random. Therefore, after optimization, we rearranged the label values such that the Dice coefficient with the ground truth segmentation was maximized. Table~\ref{tab1} shows the means and standard deviations of the Dice coefficients we obtained for the different methods on the dataset. It can be seen that the SVGMM improves segmentation accuracy compared to the GMM. However, for both GMM and SVGMM including the CNN in the optimization algorithm further improved the Dice scores. An example of the obtained segmentation masks is illustrated in Figure~\ref{fig:2}. There it can be observed that although the CNN-based methods fail to minimize the NLL exactly, this comes with the benefit of spatially more regularized segmentation masks, which increases the similarity to the ground truth masks.

\subsection{Training on multiple images}
\begin{figure}[!b]
\includegraphics[width=\textwidth]{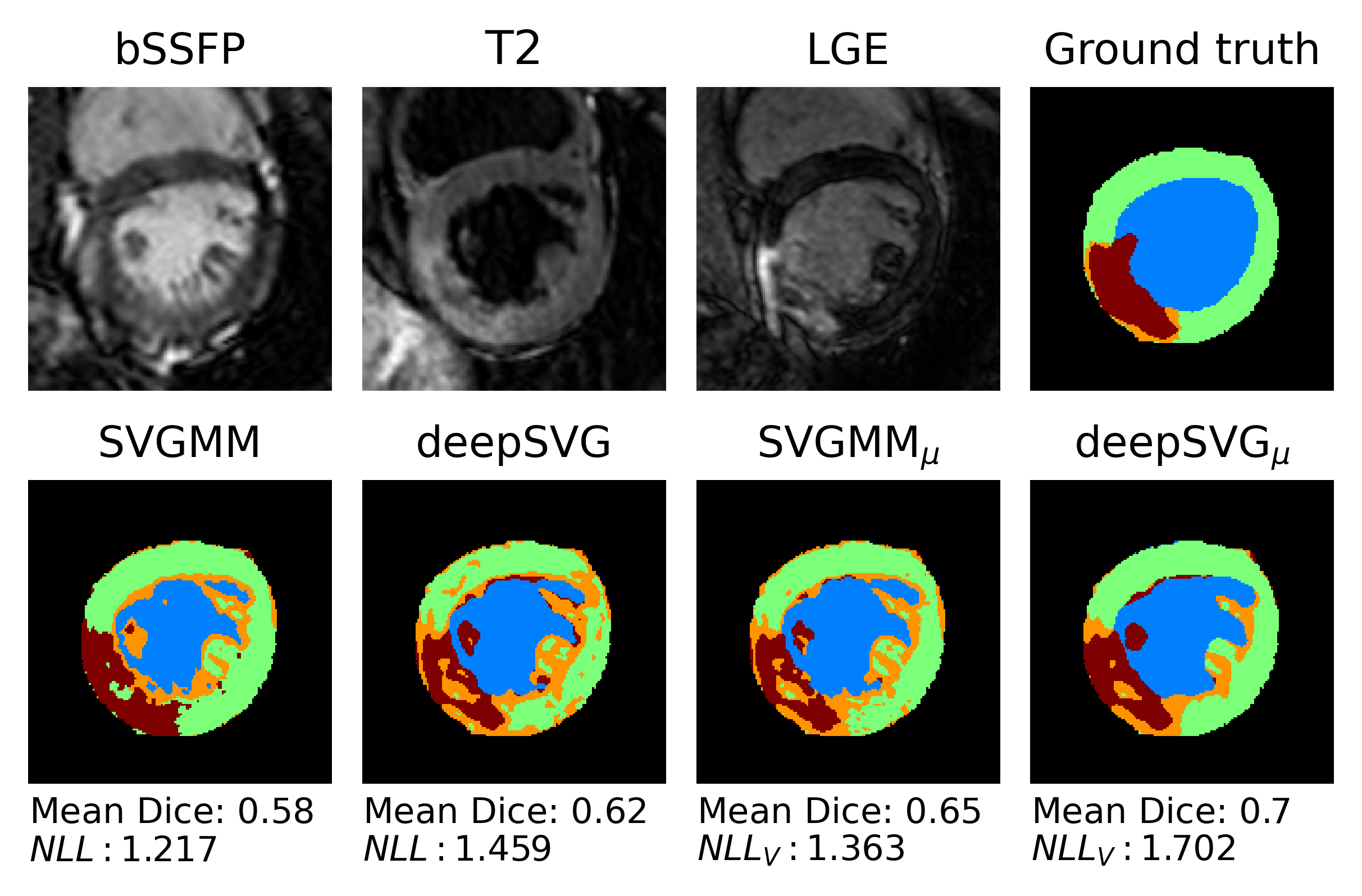}
\caption{Example of an image in the test dataset. Including mean estimation $\boldsymbol{\mu}^{\text{data}}$ into the methods improved segmentation performance.} \label{fig:3}
\end{figure}
In a second experiment we tested how well a network, which was trained on a dataset consisting of multiple images, can generalize to unseen data. To this end, we split the dataset into a training dataset ($N=60$) and a test dataset ($N=19$) and trained the deepSVG according to Algorithm \ref{alg:proposed_train}, without any regularization. Also, we investigated how adding a regularization term during training could further improve segmentation accuracy. To this end we tested a regularization term that incorporated a rough estimate for the mean values $\boldsymbol{\mu}$. For this purpose, we took $10$ random images from the training dataset and calculated the ``true'' mean values of the tissue types with the help of the ground truth labels. We the averaged those obtained values to get one single vector $\boldsymbol{\mu}^{\text{data}}$ and trained a neural network as described in Algorithm \ref{alg:proposed_train} with regularizing function
\begin{align*}
    r(\boldsymbol{\mu}) = \norm{\boldsymbol{\mu} - \boldsymbol{\mu}^{\text{data}}}_2^2
\end{align*}
and regularizing parameter $\lambda=1$. We refer to this method as deepSVG$_\mu$. We compared these two CNN-based methods with the classical SVGMM and a method we call SVGMM$_\mu$, in which we initialized the EM algorithm with $\boldsymbol{\mu}^{\text{data}}$. Means and standard deviations of the Dice coefficients for the described methods on the test dataset are shown in Table~\ref{tab:2}. It turned out that by including $\boldsymbol{\mu}^{\text{data}}$ into the framework rearranging the labels became unnecessary. Although the mean Dice scores of the labels obtained by the unregularized  CNN were slightly lower, the method's performance was not much worse than the standard SVGMM. When initialized/regularized with $\boldsymbol{\mu}^{\text{data}}$, Dice scores of both methods improved. However, adding this simple regularization term proved very effective for the deepSVG model posting the best Dice coefficients on all tissue types compared to the other methods. As Figure~\ref{fig:3} shows a single example comparing the different methods, we can again observe the phenomenon that the NLL values which are achieved by deepSVG$_\mu$ are higher than the ones obtained through SVGMM$_\mu$, indicating non-optimal NLL minimization. However, when comparing with the ground truth labels, the overall segmentation performance was significantly better for the CNN-based method.

\begingroup
\setlength{\tabcolsep}{5pt}
\begin{table}[t]
\caption{A comparison of segmentation performance for four different methods. The entries show the mean Dice coefficients $\pm$ standard deviations obtained over the $19$ images of the test dataset. The best scores are marked in bold.}\label{tab:2}
\centering
\begin{tabular}{l| c c c c c c}
\hline
Method      & blood                & muscle                & edema                 & scar                 & mean   \\
\hline \hline
SVGMM       & $0.79 \pm 0.05$      & $ 0.78 \pm 0.06$      & $ 0.09  \pm 0.12$     & $ 0.52 \pm 0.17$     & $ 0.55\pm0.05  $  \\
deepSVG         & $0.70 \pm 0.11$      & $ 0.79 \pm 0.04$      & $ 0.12  \pm 0.08$     & $ 0.49 \pm 0.16$     & $ 0.52\pm0.04$ \\
SVGMM$_\mu$ & $0.79 \pm 0.04$      & $ 0.82 \pm 0.05$      & $ 0.21  \pm 0.12$     & $ 0.49 \pm 0.28$     & $ 0.58\pm0.10  $\\
deepSVG$_\mu$   & $\bm{0.86 \pm 0.04}$ & $ \bm{0.83 \pm 0.06}$ & $\bm{0.29  \pm 0.15}$ & $\bm{0.53 \pm 0.25}$ & $\bm{0.63 \pm0.08} $\\
\hline
\end{tabular}
\end{table}
\endgroup

\subsection{Implementation details}

All the experiments have been implemented in Python using PyTorch \cite{NEURIPS2019_bdbca288} for the deep learning parts. For the GMM the implementation of the scikit-learn package \cite{JMLR:v12:pedregosa11a} was used. In all of our experiments we restricted the covariance matrices $\Sigma$ to be of diagonal form. However, extending the proposed method to more general covariance matrices would be straightforward. For all the trained CNNs the U-net architecture \cite{ronneberger2015u} was used and for the gradient-step the AdamW \cite{loshchilov2017decoupled} algorithm was applied with a learning rate of $0.001$. It is worth mentioning that in our experiments we did not calculate the full gradients in each step as described in Algorithms \ref{alg:proposed} and \ref{alg:proposed_train}, but only used some gradient approximation. Before applying the different methods, all the images in the dataset were normalized to have a mean of zero and a standard deviation of one. The full implementation code of this project will be publicly available at \url{https://github.com/matthi99/DeepGMM.git}.

\section{Conclusion}
\label{sec:conclusion}

In this work we developed a novel method which is combining the ML estimation of mixture models with deep learning techniques. We explain how GMMs and SVGMMs can be used for multi-source image segmentation and point out their main weakness, which is assuming statistical independence between all the pixel values. Inspired by the structure of the EM algorithm, we replaced the expectation step in the algorithm by a deep learning gradient step. This way we were able to get ML estimates for both the GMM and the SVGMM using CNNs. However, due to the convolutional structure of the CNNs we implicitly achieve spatial regularization for the segmentation masks. By performing experiments for myocardial image pathology segmentation, we demonstrate that our method is applicable for real-life problems. We show that when optimizing on one single image, our CNN-based methods outperforms the classical methods in terms of segmentation accuracy. By training a deep learning network on a training dataset we also show the generalization potential of our framework, making ML optimization for every single image redundant. Lastly, we demonstrate the flexibility of our method regarding additional regularization and demonstrate big performance improvements by adding a fairly simple regularization term. Future work could focus on further improving segmentation accuracy of the method by exploring new regularization possibilities. This could include extending the framework to semi-supervised applications, where for a small part of the training images ground truth labels are available. Finally, the method could be extended to model intensity distributions not only with Gaussian distributions, but also with more complex distributions.

\begin{credits}
\subsubsection{\ackname} This work was supported by the Austrian Science Fund (FWF) [grant number DOC 110].

\subsubsection{\discintname}
The authors have no competing interests to declare that are
relevant to the content of this article.
\end{credits}
%
% ---- Bibliography ----
%
% BibTeX users should specify bibliography style 'splncs04'.
% References will then be sorted and formatted in the correct style.
%
\bibliographystyle{splncs04}
\bibliography{references}
%

%\begin{thebibliography}{8}
%\bibitem{ref_article1}
%Author, F.: Article title. Journal \textbf{2}(5), 99--110 (2016)

%\bibitem{ref_lncs1}
%Author, F., Author, S.: Title of a proceedings paper. In: Editor,
%F., Editor, S. (eds.) CONFERENCE 2016, LNCS, vol. 9999, pp. 1--13.
%Springer, Heidelberg (2016). \doi{10.10007/1234567890}

%\bibitem{ref_book1}
%Author, F., Author, S., Author, T.: Book title. 2nd edn. Publisher,
%Location (1999)

%\bibitem{ref_proc1}
%Author, A.-B.: Contribution title. In: 9th International Proceedings
%on Proceedings, pp. 1--2. Publisher, Location (2010)

%\bibitem{ref_url1}
%LNCS Homepage, \url{http://www.springer.com/lncs}, last accessed 2023/10/25
%\end{thebibliography}
\end{document}